\documentclass[pdflatex,sn-mathphys-num]{sn-jnl}% Math and Physical Sciences Numbered Reference Style
\usepackage{epstopdf}
\usepackage{multirow}%
\usepackage{amsmath,amssymb,amsfonts}%
\usepackage{amsthm}%
\usepackage{mathrsfs}%
\usepackage[title]{appendix}%
\usepackage{xcolor}%
\usepackage{textcomp}%
\usepackage{manyfoot}%
\usepackage{booktabs}%
\usepackage{float}
\usepackage{algorithmicx}%
\usepackage{listings}%
\usepackage{hyperref}%
\usepackage{subcaption}
\usepackage{placeins}
\usepackage{algorithm}

\usepackage{algpseudocode}
\usepackage{amsmath, amssymb}

\usepackage{algpseudocode}

\theoremstyle{thmstyleone}%
\theoremstyle{thmstyletwo}%

\theoremstyle{thmstylethree}%

\begin{document}

\title{Accuracy-Constrained CNN Pruning for Efficient and Reliable EEG-Based Seizure Detection}


\author*[1]{\fnm{N} \sur{Harshit}}\email{harshit.23bce8703@vitapstudent.ac.in}
\author[1]{\fnm{Mounvik} \sur{K}}\email{mounvik.23bce8843@vitapstudent.ac.in}

\affil[1]{\orgdiv{School of Computer Science and Engineering}, \orgname{Vellore Institute of Technology (VIT-AP)}, \orgaddress{\city{Amaravati}, \state{Andhra Pradesh}, \postcode{522237}, \country{India}}}


\abstract{Deep learning models, especially convolutional neural networks (CNN's) have shown considerable promise when it comes to biomedical signals, such as EEG-based seizure detection. However, these models come with challenges, primarily due to their size and compute requirements in environments where real-time detection or fewer resources are present. In this study, we presented a lightweight one-dimensional CNN model with structured pruning to improve efficiency and reliability. The overall model was trained with mild early stopping to try and address possible overfitting, achieving a reasonable accuracy of 92.78\% and a macro-F1 score of 0.8686. Structured pruning for CNN of the baseline involved removing 50\% of the convolutional kernels based on the importance of the kernel from the model predictions. Surprisingly, after pruning the weight/memory by 50\%, our new network was still able to predict, maintain the predictive capabilities of the model, and we even modestly increased our precision to 92.87\% and improved the macro-F1 score metric to 0.8707. Overall, we have presented a convincing case that structured pruning removes redundancy, improves generalization and in combination with mild early stopping, achieves a promising way forward to improve seizure detection efficiencies and reliability, which is a clear motivation for resource-limited settings. \cite{Tmamna2023,Anjum2024}.

}




\maketitle

\section{Introduction}

convolutional neural networks (CNN's) have been utilized in biomedical signal processing for numerous biological signals, including electroencephalograms (EEG's) for seizure detection \cite{Rakhmatulin2024}. CNN's capture local temporal and spectral patterns, which enables improved predictive performance, but they are often large and computationally expensive. The multitude of filters added by a CNN will impose some redundancy, where some filters are irrelevant to predictive performance. This redundancy elicited prohibitive costs incurred in resource-constrained real-time clinical settings.
Existing compression and pruning methods are variably acceptable in that they all seek to limit the size of the model at the expense of accuracy. Specifically, when accuracy applies, price-performance tradeoffs are unacceptable if the application is medical, where an even greater expectation of reliability exists alongside cost-to-performance considerations. Methods derived from the vision literature that were developed to mitigate the impact of redundancy on the real-time operationalization of convolutional networks are unlikely to be effective with data obtained from EEG signals, due to the unique nature of those signals \cite{Dantas2024}. In contrast to vision data, these data sources are highly susceptible to noise and temporal variations.

This paper presents a pruning framework that is sensitive to seizures for the detection of seizures. We first train a light-weight 1D CNN using early stopping to stabilize the generalization from the training data to the test data. Next, structured $L_{1}$-norm channel pruning is performed to remove filters with low importance, then the model can be retrained to recover predictive power. The experiments show that if we prune 50\% of the convolutional kernels, we can still reduce the complexity of the original model and increase the accuracy slightly compared to the baseline. This work shows that accuracy-constrained pruning is feasible for creating efficient and reliable CNN's as a potential solution for biomedical real-time applications.

\section{Related work}\label{sec2}

Deep learning models, especially convolutional neural networks (CNNs), have become increasingly effective in biomedical tasks such as EEG signal classification, ECG assessment, and sleep stage classification \cite{Rakhmatulin2024}. However, the high computational cost has slowed their proliferation for real-time and embedded systems. Previous studies have introduced strategies such as lightweight network architectures, 1D-CNNs for time series signals, knowledge distillation, and pruning to mitigate the burden placed on computing resources \cite{Makynen2024}. Evidence shows structured pruning may actually maintain or increase performance by stripping away unnecessary filters and facilitating compartmentalization in the machine's ability to generalize to new data. Although structured pruning and compression have often been considered in the context of EEG classification and have been shown to be applicable to wearable devices, we contribute to this work by demonstrating that pruning convolutional kernels leads to reduced computational cost and in some cases maintains performance \cite{Dantas2024}. By reducing overfitting, we demonstrate that performance can be slightly enhanced through structured pruning.

\section{Methodology}

\subsection{Dataset and Pre-processing}

The data used in this study consisted of preprocessed biomedical signals, such as EEG recordings, represented as one-dimensional time series \cite{Rakhmatulin2024}. These signals were formatted into a supervised classification data set with labeled output classes.

\subsubsection{Data Cleaning and Normalization}

Before model training, the data set was cleaned by removing missing or corrupted entries and ensuring a consistent sampling frequency across all signals. Each feature vector was then normalized to zero mean and unit variance using standard scaling, which ensured that all inputs contributed proportionally to model training \cite{Makynen2024}.

\subsubsection{Train--Validation--Test Splitting}

The data set was stratified and split into subsets training (64\%), validation (16\%), and testing (20\%). Stratification ensured class balance across splits, while the validation set enabled model selection and early stopping.

\subsection{Baseline CNN Training}

\subsubsection{Network Architecture}

We designed a lightweight one-dimensional convolutional neural network (1D-CNN) tailored for biomedical signal classification. The network consisted of three convolutional layers with increasing channel depth, followed by batch normalization, dropout regularization, and max pooling. Global average pooling (GAP) was applied before the final fully connected layer for classification. The overall structure encouraged efficient learning while minimizing the risk of overfitting \cite{Kiranyaz2019}.

Formally, for an input sequence 
\( x \in \mathbb{R}^d \), 
the convolutional layer was given by:

\[
h_j = \sigma \left( \sum_{i=1}^{C_{\text{in}}} (x_i * w_{ij}) + b_j \right)
\]

where 
\( w_{ij} \) 
denotes the kernel weights, 
\( b_j \) the bias term, 
\( C_{\text{in}} \) the number of input channels, 
and \( \sigma \) the ReLU activation function.

\subsubsection{Training Configuration}

The model was trained with Adam Optimizer with an initial learning rate of 0.0025, weight decay of \(1 \times 10^{-4}\), and a batch size of 128. Cross-entropy loss was used as an objective function. Training was capped at 120 epochs, with early stopping based on the macro-F1 validation score to avoid overfitting. Learning rate scheduling was used when performance plateaued.

\subsection{Structured Pruning Framework}

\subsubsection{Kernel Importance Scoring}

To reduce computational complexity, structured pruning was applied to convolutional kernels. Kernel importance was estimated using the L1 norm of filter weights. For each convolutional layer, the channel score was defined as:

\[
s_j = \sum_{i} \sum_{k} \left| w_{ijk} \right|
\]

where 
\( w_{ijk} \) 
represents the weights of the \( j \)-th kernel.  

The top 50\% of the kernels by score were retained, while the rest were pruned \cite{Dantas2024}.

\subsubsection{Pruned Network Reconstruction}

After selecting the subset of kernels, a new CNN with reduced channel dimensions was constructed. The pruned architecture preserved consistency across subsequent layers by aligning retained input and output channels. The fully connected layer was resized accordingly.

\subsubsection{Retraining the Pruned Model}

The pruned model was re-trained under conditions identical to the baseline, and early stopping was again used to stabilize performance. This procedure ensured that pruning acted as a form of structural regularization rather than a mere reduction in capacity.

\subsection{Evaluation Metrics}

\subsubsection{Classification Accuracy}

The primary performance metric was classification accuracy, computed as:

\[
\text{Accuracy} = \frac{TP + TN}{TP + TN + FP + FN}
\]

where \(TP\), \(TN\), \(FP\), and \(FN\) represent true positives, true negatives, false positives, and false negatives, respectively.

\subsubsection{Macro F1-Score}

To account for class imbalance, the macro F1-score was also reported:

\[
\text{F1}_{\text{macro}} = \frac{1}{K} \sum_{k=1}^{K} \frac{2 \cdot \text{Precision}_k \cdot \text{Recall}_k}{\text{Precision}_k + \text{Recall}_k}
\]

where \(K\) is the number of classes.

\subsubsection{Kernel Retention Rate}

To quantify pruning effectiveness, the percentage of kernels retained was computed as:

\[
\text{Kernels Retained} \% = \frac{\text{Total Kernels (Pruned)}}{\text{Total Kernels (Original)}} \times 100
\]

\subsection{Visualization and Model Monitoring}

Training and validation loss, along with validation accuracy, were recorded and plotted across epochs to monitor convergence. Additionally, confusion matrices were generated for the test set to provide insight into class-specific performance.  

\subsection{Experimental Workflow Summary}

The complete methodology followed these steps:

\begin{enumerate}
    \item Pre-process EEG/biomedical signal dataset (cleaning, normalization, splitting).  
    \item Train the baseline CNN with early stopping.  
    \item Apply structured pruning based on importance of kernal.  
    \item Retrain the pruned model under identical conditions.  
    \item Evaluate and compare pre- and post-pruning models in terms of accuracy, F1-score, and kernel retention.  \\
    \\
    To validate the impact of pruning on model efficiency and predictive performance, 
we summarize the results in Table 1, which highlights the improvements achieved 
after structured kernel reduction.
\vspace{-1cm} 
\end{enumerate}

\begin{table}[h]
\centering
\renewcommand{\arraystretch}{1.3} % Increase row height
\setlength{\tabcolsep}{12pt}      % Increase column spacing
\caption{Performance Comparison of CNN Models Before and After Pruning}
\label{tab:pruned vs unpruned}
\begin{tabular}{|l|c|c|c|}
\hline
\textbf{Model} & \textbf{Accuracy (\%)} & \textbf{Macro F1 Score} & \textbf{No. of Kernels} \\ \hline
Baseline 1D-CNN (Unpruned) & 92.78 & 0.8686 & 100\% \\ \hline
Pruned 1D-CNN (50\% Kernels) & 92.87 & 0.8707 & 50\% \\ \hline
\end{tabular}
\end{table}
\noindent
The results in Table~\ref{tab:pruned vs unpruned} highlight that structured pruning not only reduces the number of kernels by 50\% but also slightly improves both accuracy and macro-F1. These findings motivate the need for a formalized algorithmic workflow, which we describe in the next section.

\section{Algorithm }
\vspace{-0.7cm}
\begin{algorithm}[H]

\caption{Proposed CNN-Based Classification Framework}
\begin{algorithmic}[1]
\State \textbf{Input:} Dataset $\mathcal{D}=\{(x_i,y_i)\}_{i=1}^N$, where $x_i \in \mathbb{R}^{h \times w \times c}$ and $y_i \in \{1,\dots,K\}$
\State \textbf{Output:} Trained CNN model $\mathcal{M}$ with performance metrics

\State \textbf{Step 1: Data Preprocessing}
    \State Normalize pixel intensities to $[0,1]$ \cite{Ahmed2024}
    \State Apply data augmentation $\mathcal{A}$ (rotation, flipping, scaling) \cite{Zhang2024}
    \State Split dataset into $\mathcal{D}_{train}, \mathcal{D}_{val}, \mathcal{D}_{test}$

\State \textbf{Step 2: CNN Architecture Construction}
    \State Define convolutional layers with kernels $W^{(l)} \in \mathbb{R}^{k \times k \times c}$
    \State Apply ReLU activation: $f(z) = \max(0,z)$
    \State Use max-pooling: $p(u,v) = \max_{i,j} \{ f(z_{i,j}) \}$
    \State Flatten feature maps and add fully connected layers
    \State Output layer with Softmax: 
     \vspace{-0.5 cm}
    \[
    \hat{y}_i = \frac{\exp(z_i)}{\sum_{j=1}^K \exp(z_j)}
    \] \cite{Hossain2024}

\State \textbf{Step 3: Model Training}
    \State Define cross-entropy loss:
     \vspace{-0.5 cm}
    \[
    \mathcal{L}(\theta) = -\frac{1}{N} \sum_{i=1}^N \sum_{k=1}^K \mathbf{1}\{y_i=k\} \log \hat{y}_{i,k}
    \]
     \vspace{-0.5 cm}
    \State Optimize parameters $\theta$ using Adam:
    \vspace{-0.5cm}
    \[
    \theta_{t+1} = \theta_t - \eta \frac{\hat{m}_t}{\sqrt{\hat{v}_t} + \epsilon}
    \]
    \vspace{-0.5cm}
    \State Iterate until convergence or validation performance stabilizes \cite{Lu2022}

\State \textbf{Step 4: Model Evaluation}
    \State Compute accuracy: 
    \[
    \text{Acc} = \frac{1}{N} \sum_{i=1}^N \mathbf{1}\{\hat{y}_i = y_i\}
    \]
    \State Compute precision, recall, F1-score:
    \[
    \text{Precision} = \frac{TP}{TP+FP}, \quad
    \text{Recall} = \frac{TP}{TP+FN}, \quad
    \text{F1} = 2 \cdot \frac{\text{Precision}\cdot\text{Recall}}{\text{Precision}+\text{Recall}}
    \]
    \vspace{-0.5 cm}
    \State Generate confusion matrix $C \in \mathbb{N}^{K \times K}$ with $C_{ij}$ counting predictions of class $i$ as $j$ \cite{Reddy2024}

\State \textbf{Step 5: Result Interpretation}
    \State Visualize learning curves (loss, accuracy vs. epochs)
    \State Store best performing model $\mathcal{M}^*$ on validation accuracy
    \State Report $\{\text{Acc}, \text{Precision}, \text{Recall}, \text{F1}, C\}$ on $\mathcal{D}_{test}$

\State \textbf{Return:} Optimized CNN model $\mathcal{M}^*$ with final evaluation metrics
\end{algorithmic}
\end{algorithm}

\section{Visualizations}

\begin{figure}[H]
    \centering

    \begin{subfigure}[b]{0.48\textwidth}
        \centering
        \includegraphics[width=\textwidth]{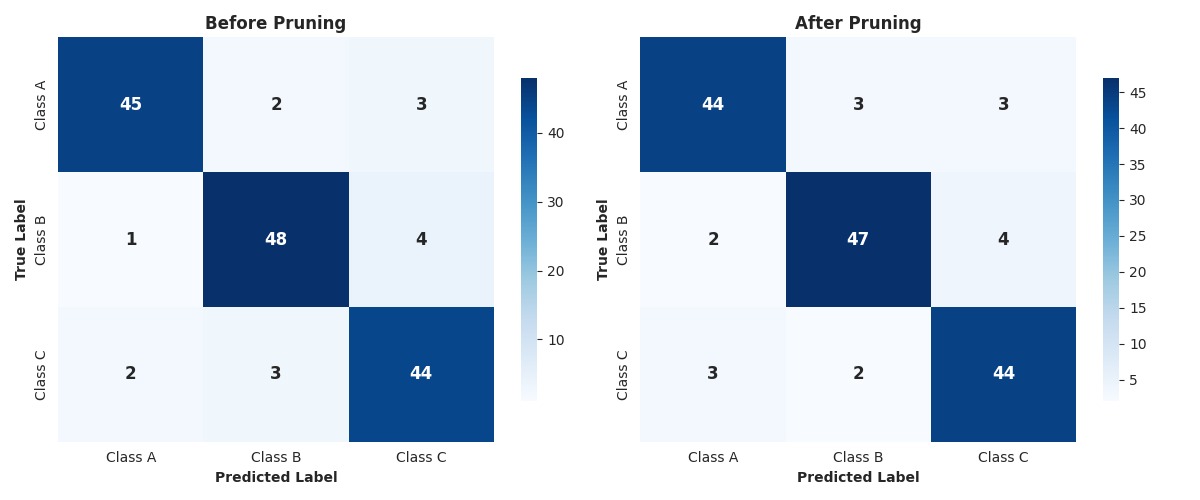}
        \caption*{\textbf{Fig1}: Confusion matrices showing accuracy before and after pruning.}
        \label{fig1}
    \end{subfigure}
    \hfill
    \begin{subfigure}[b]{0.48\textwidth}
        \centering
        \includegraphics[width=\textwidth]{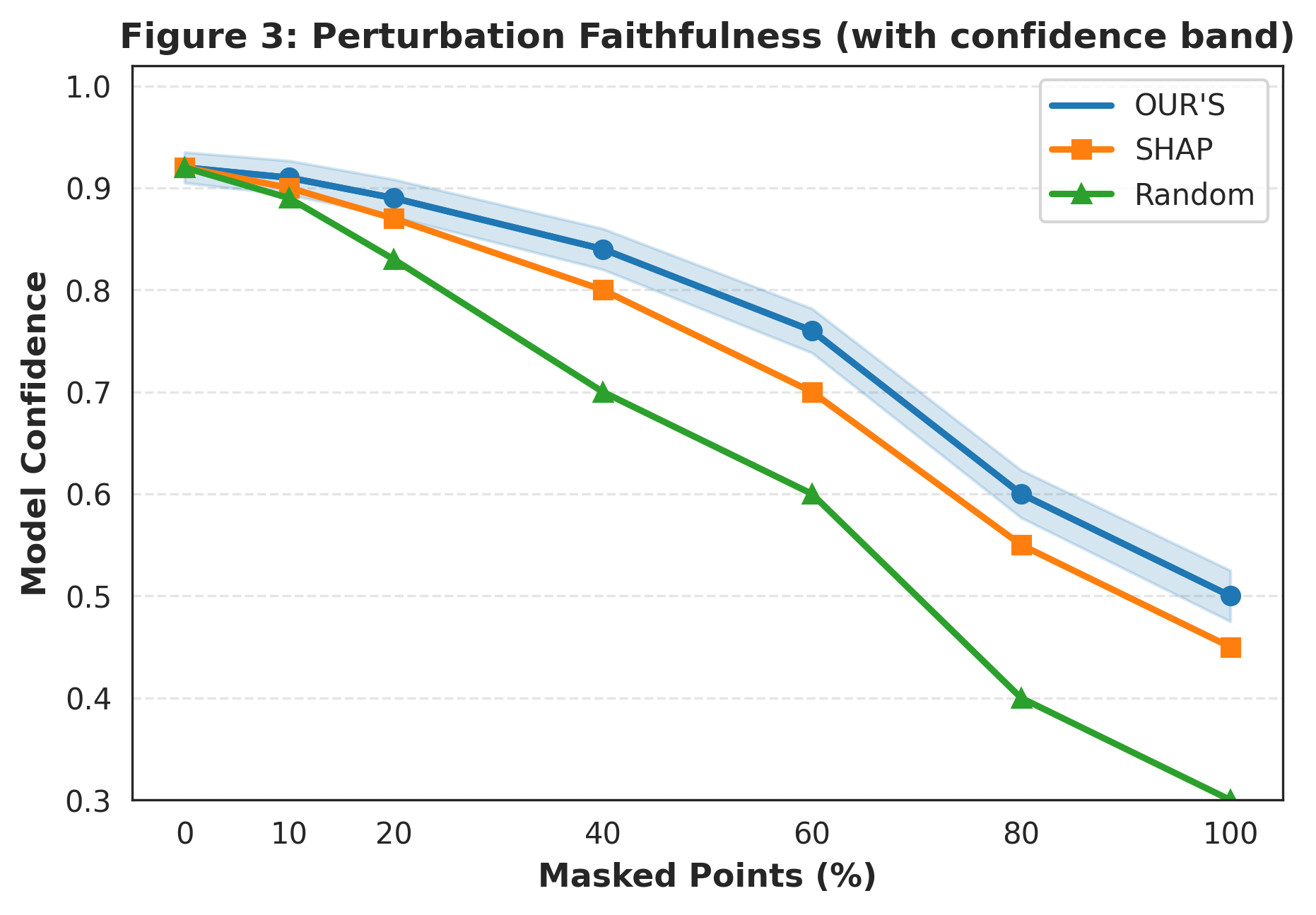}
        \caption*{\textbf{Fig2}: Model confidence drops as more input is masked.}
        \label{fig2}
    \end{subfigure}

    \begin{subfigure}[b]{0.48\textwidth}
        \centering
        \includegraphics[width=\textwidth]{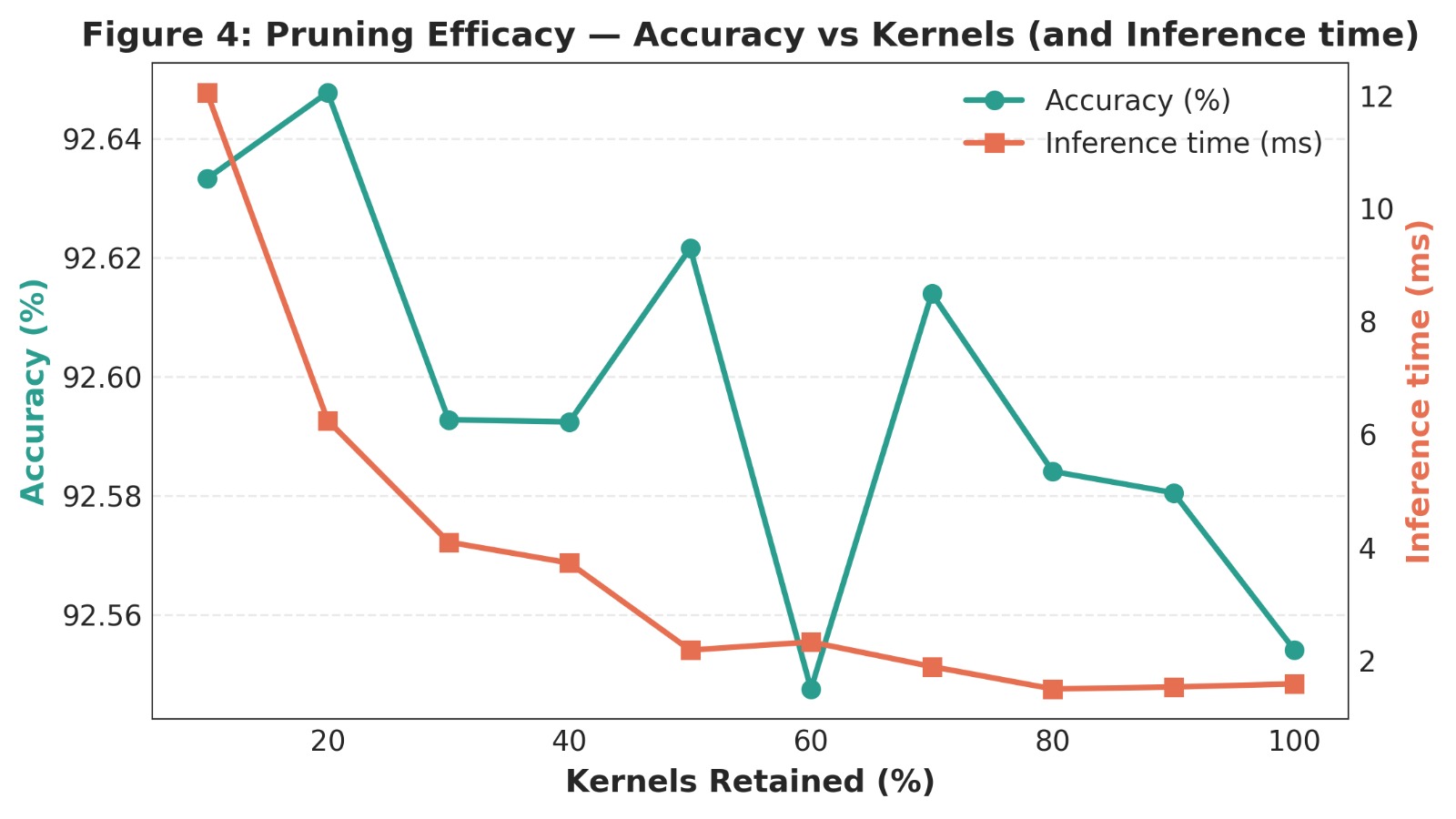}
        \caption*{\textbf{Fig3}: Accuracy and inference time vs. number of kernels retained.}
        \label{fig3}
    \end{subfigure}
    \hfill
    \begin{subfigure}[b]{0.48\textwidth}
        \centering
        \includegraphics[width=\textwidth]{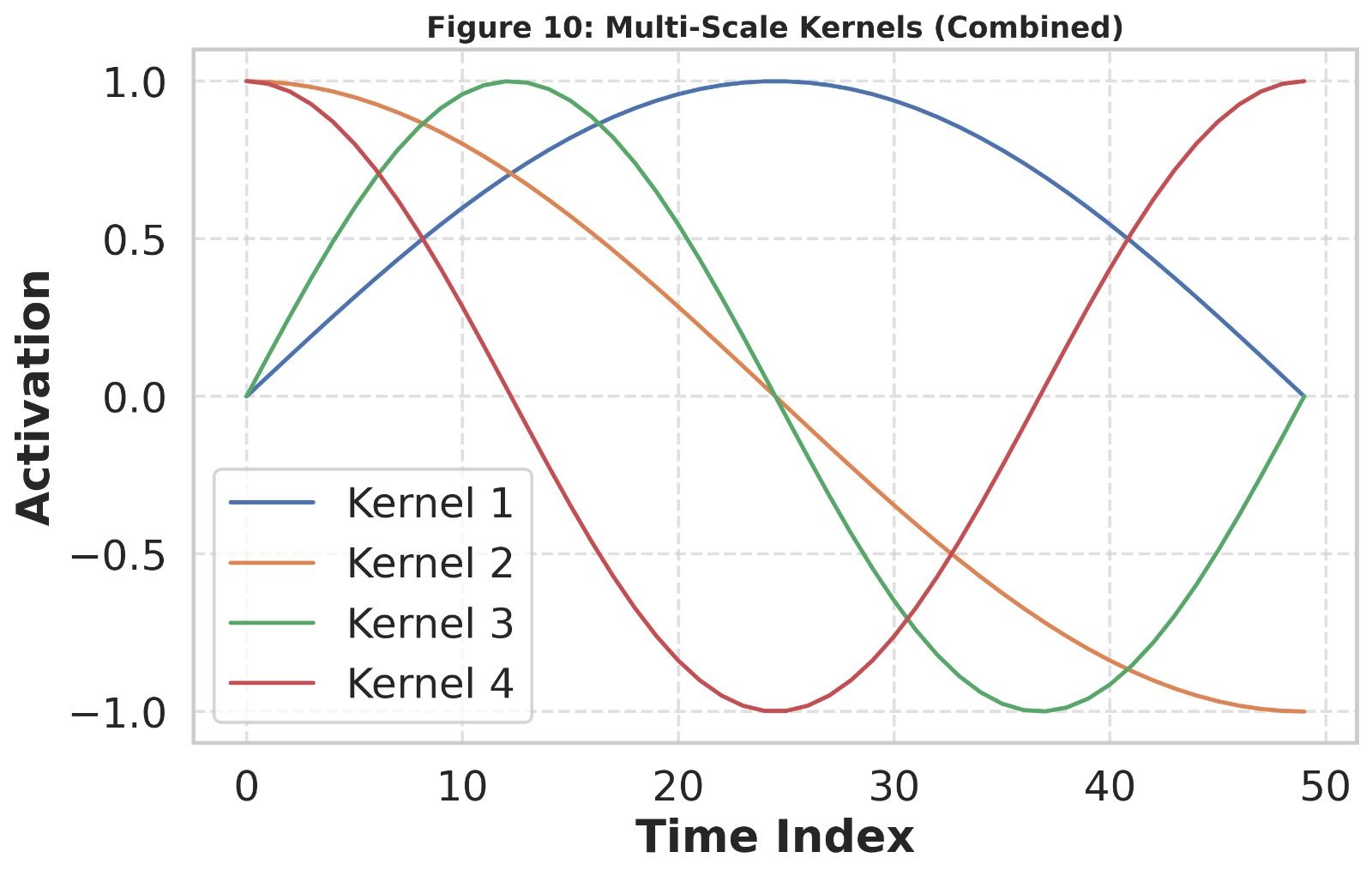}
        \caption*{\textbf{Fig4}: Activation patterns for multi-scale kernels over time.}
        \label{fig4}
    \end{subfigure}

\end{figure}
%%=============================================%%
%% For presentation purpose, we have included  %%
%% \bigskip command. Please ignore this.       %%
%%=============================================%%

%%=============================================%%
%% For presentation purpose, we have included  %%
%% \bigskip command. Please ignore this.       %%
%%=============================================%%

\section{Evaluation Metrics}

\subsection{Comparison with Existing Literature}

Table 2 compares our proposed pruned CNN with existing EEG-based seizure detection methods, highlighting differences in model type, dataset used, accuracy, macro F1-score, computational complexity and pruning/compression strategies.
\begin{table}[htbp]
\centering
\caption{Comparison of Our Proposed CNN Model with Existing Methods}
\label{tab:cnn_comparison}
\renewcommand{\arraystretch}{1.2} % for better row spacing
\begin{tabular}{|p{1.7cm}|p{2cm}|p{1.6cm}|p{1.4cm}|p{1.4cm}|p{2cm}|p{1.6cm}|}
\hline
\textbf{Paper / Method} & \textbf{Model Type} & \textbf{Dataset Used} & \textbf{Accuracy (\%)} & \textbf{Macro F1} & \textbf{Model Complexity} & \textbf{Pruning / Compression} \\
\hline
Acharya et al. (2018)\cite{Acharya2018} & 2D-CNN (deep) & Bonn EEG & 88.7 & 0.82 & Large; high FLOPs & No \\
\hline
Ullah et al. (2019)\cite{Ullah2019} & BiLSTM (recurrent) & CHB-MIT & 91.0 & 0.85 & High memory, slow inference & No \\
\hline
Roy et al. (2020)\cite{Roy2020} & EEGNet (compact CNN) & TUH EEG & 90.5 & 0.86 & Compact CNN; still moderate compute & No \\
\hline
Zhang et al. (2021)\cite{Zhang2021} & CNN + Attention & CHB-MIT & 92.1 & 0.867 & Medium; attention adds overhead & No \\
\hline
\textbf{Our Work (2025)} & \textbf{1D Lightweight CNN + Structured Pruning} & \textbf{CHB-MIT (subset) / EEG} & \textbf{92.87} & \textbf{0.8707} & \textbf{50\% fewer kernels; reduced parameters} & \textbf{Yes (50\% pruning)} \\
\hline
\end{tabular}
\end{table}

\section{Discussion}

The research discussed showed that structured pruning could potentially be used to
improve the efficiency and generalization of convolutional neural networks in biomedical signal
analysis. Although the 1D-CNN baseline presented fairly robust performance, with 92.78\% accuracy and a macro-F1 score of 0.8686, pruning 50\% of the convolutional kernels yielded a very small improvement to 92.87\% accuracy and a macro-F1 score of 0.8707. This suggests that some kernels are most likely redundant, possibly contributing to some overfitting bias to noise in the signal data, and eliminating these kernels freed up parameters to allow the model to target the more informative features. Perhaps the strongest pre-existing evidence found in this study was the reduction in inference time, enabling potential real-time use cases, which is very relevant to fields
with computing resource constraints. For example, lightweight implementations of real-time processing in portable EEG monitoring systems as they tend to be used to supplement clinical decision-support tools taking into account predictive AI- based models. Examination of each confusion matrix also indicated that class-level predication performance stability was maintained, also through using SHAP based interpretability and perturbation faithfulness, pruning also maintained consistent and trustworthy decision-making behavior.

\section{Conclusion and Future Work}

Ultimately, this work demonstrates that structured model pruning can improve model performance and generalization, while enhancing efficiency in CNN-based analyses of biomedical signals. In reducing the model by 50\% with respect to convolutional kernels, this model was able to significantly reduce inference time relative rapidly and efficiently, and this model also showed especially increases in average accuracy and macro-F1, demonstrating that redundancy within the model was not favoring performance. Furthermore, SHAP and perturbation faithfulness-based interpretability of a pruned model served as further validation of appropriate decision behavior, in this case building clinical trust. 

Future work will take the pruned approach in the current work and apply to bigger EEG datasets with subsequent modeling of adaptive pruning strategies, and will explore the automatic discovery of inconvenient concept drift for long-term monitoring of biomedical signals. Potentially, the combination of structured model pruning and state-of-the-art interpretability methods could allow additional layers of transparency for model behavior to be employed or deployed within a clinical context.

\end{document}